\documentclass[journal]{IEEEtran}
\usepackage{amsmath}
\usepackage{balance}
\usepackage{graphicx}
\usepackage[hidelinks]{hyperref}
\usepackage{multirow} 
\usepackage{orcidlink}
\usepackage{subfigure}

\begin{document}
\bstctlcite{IEEEexample:BSTcontrol}

\title{Knowledge-Aware Modeling with Frequency Adaptive Learning for Battery Health Prognostics}

\author{Vijay Babu Pamshetti\raisebox{0.5ex}{\orcidlink{0000-0003-4850-5369}},~\IEEEmembership{Member,~IEEE,}
Wei Zhang\raisebox{0.5ex}{\orcidlink{0000-0002-2644-2582}},~\IEEEmembership{Member,~IEEE,}
Sumei Sun\raisebox{0.5ex}{\orcidlink{0000-0002-1701-8122}}, \IEEEmembership{Fellow, IEEE,}\\
Jie Zhang\raisebox{0.5ex}{\orcidlink{0000-0001-8996-7581}},
Yonggang Wen\raisebox{0.5ex}{\orcidlink{0000-0002-2751-5114}}, \IEEEmembership{Fellow,~IEEE}
and Qingyu Yan\raisebox{0.5ex}{\orcidlink{0000-0003-0317-3225}} 
\thanks{Manuscript received January 1, 2025; revised January 1, 2025; accepted January 1, 2025. This research is supported by A*STAR under its MTC Programmatic (Award M23L9b0052), MTC Individual Research Grants (IRG) (Award M23M6c0113), and SIT’s Ignition Grant (STEM) (Grant ID: IG (S) 2/2023 – 792). (\textit{Corresponding author: Wei Zhang})}
\thanks{Vijay Babu Pamshetti is with both the Information and Communications Technology Cluster, Singapore Institute of Technology, Singapore 828608, and the Department of EEE, Chaitanya Bharathi Institute of Technology, India 500075  (e-mails: vijaybabu.pamshetti@singaporetech.edu.sg and vijaybabup eee@cbit.ac.in)}
\thanks{Wei Zhang and Sumei Sun are with the Information and Communications Technology Cluster, Singapore Institute of Technology, Singapore 828608 (e-mails: wei.zhang@singaporetech.edu.sg and sumei.sun@singaporetech.edu.sg).}
\thanks{Yonggang Wen and Jie Zhang are with the College of Computing and Data Science, Nanyang Technological University, Singapore 639798 (e-mail: ygwen@ntu.edu.sg and zhangj@ntu.edu.sg).}
\thanks{Qingyu Yan is with the School of Materials Science and Engineering, Nanyang Technological University, Singapore 639798 (e-mail: alexyan@ntu.edu.sg).}
}

%
%

\markboth{IEEE Transactions on xxx,~Vol.~1, No.~1, January~2025}%
{Shell \MakeLowercase{\textit{et al.}}: Bare Demo of IEEEtran.cls for IEEE Journals}

\maketitle

\begin{abstract}
Battery health prognostics are critical for ensuring safety, efficiency and sustainability in modern energy systems. However, it has been challenging to achieve accurate and robust prognostics due to complex battery degradation behaviors with nonlinearity, noises, capacity regeneration, etc. Existing data-driven models capture temporal degradation features but often lack knowledge guidance, which leads to unreliable long-term health prognostics. To overcome these limitations, we propose \textsc{Karma}, a \underline{k}nowledge-\underline{a}wa\underline{r}e \underline{m}odel with frequency-\underline{a}daptive learning for battery capacity estimation and remaining useful life prediction. The model first performs signal decomposition to derive battery signals in different frequency bands. A dual-stream deep learning architecture is developed, where one stream captures long-term low-frequency degradation trends and the other models high-frequency short-term dynamics. \textsc{Karma} regulates the prognostics with knowledge, where battery degradation is modeled as a double exponential function based on empirical studies. Our dual-stream model is used to optimize the parameters of the knowledge with particle filters to ensure physically consistent and reliable prognostics and uncertainty quantification. Experimental study demonstrates \textsc{Karma}’s superior performance, achieving average error reductions of 50.6\% and 32.6\% over state-of-the-art algorithms for battery health prediction on two mainstream datasets, respectively. These results highlight \textsc{Karma}’s robustness, generalizability and potential for safer and reliable battery management across diverse applications.
\end{abstract}


\IEEEpeerreviewmaketitle

\section{Introduction}
\IEEEPARstart{L}{i-ion} batteries have become essential for modern energy systems, powering applications from consumer electronics to electric vehicles (EVs) \cite{zhang2024role} and renewable energy storage \cite{qi2024joint}. With the global demand expected to exceed 4.7 TWh by 2030, ensuring safe and efficient operation through accurate battery health prognostics is increasingly critical \cite{fleischmann2023battery}. Reliable prognostics not only improve battery safety and performance but also reduce lifecycle costs and support sustainable energy transitions. However, complex electrochemical processes, nonlinear degradation behaviors and capacity regeneration phenomena present challenges in achieving accurate and robust health prognostics, and the industry demands advanced prognostics methodologies \cite{Zhao2024Practical}.

Two common prognostic problems include the state of health (SoH) estimation and remaining useful life (RUL) prediction. SoH as a percentage is quantified as the ratio of remaining maximum capacity to rated capacity, and RUL is the number of charging/discharging cycles from the current cycle to the cycle of a battery's end-of-life (EoL), e.g., when a battery's SoH drops to 70\%. Despite years of research, accurate and robust battery health prognostics remain challenging. The research has progressed through several distinct paradigms. Each addresses specific limitations while introducing new challenges requiring further improved solutions.

Pure data-driven and machine learning (ML) based approaches established modern battery prognostics foundations. Zhang et al. \cite{zhang2018long} demonstrated the capability of the long short-term memory (LSTM) in capturing complex temporal dependencies within degradation data. Li et al. \cite{li2023remaining} advanced the field by integrating temporal convolutional network (TCN), gated recurrent unit (GRU) and deep neural network (DNN) with dual attention mechanisms to achieve improved accuracy and efficiency. Another hybrid solution with convolutional neural network (CNN), LSTM and DNN has been presented in \cite{zraibi2021remaining} for RUL estimation. Anh et al. \cite{anh2024prediction} enhanced temporal modeling through a framework with CNN and bidirectional LSTM (BiLSTM) with attention mechanisms for both SoH and RUL predictions. Transformer-based models with attention mechanisms have also been developed for battery prognostics. Wang et al. \cite{wang2023remaining} introduced dual-branch architectures with domain-specific biases, and an exponential Transformer model was proposed in \cite{wang2025exponential} with decay strategies for temporally proximate features. Park et al. \cite{park2025detailed} leveraged multi-head self-attention for battery capacity prediction. Existing research works also emphasized feature integration, with Yao et al. \cite{yao2025remaining} combining statistical features with physical discharge features, and Zhang et al. \cite{10253731} developing specialized self-attention graph pooling networks. 


Despite remarkable progress in modeling temporal dependencies and nonlinear relationships, data-driven approaches have limitations. These approaches struggle to understand degradation patterns comprehensively and achieve optimal performance, especially for long-term prognostics. To address the limitations, researchers have increasingly adopted signal processing to augment the time-series based degradation modelling. Liu et al. \cite{9040661} focused on decomposing battery signals using empirical mode decomposition (EMD) into different frequency bands and then used LSTM to analyze the decomposed signals. Duan et al. \cite{duan2025lithium} advanced the concept by incorporating adaptive noise-based EMD and implementing parallel BiLSTM with multi-head attention mechanisms. Variational mode decomposition (VMD) subsequently emerged as a more effective approach \cite{yuan2024improved}, and was applied for battery prognostics for its superior noise resistance and strong mathematical foundations \cite{6655981}. Building on VMD's capabilities, Ding et al. \cite{9758685} developed integrated frameworks combining VMD with GRUs through meta-heuristic optimization strategies, and Chen et al. \cite{CHEN2024113388} explored a VMD-Transformer hybrid architecture. Addressing the inherent complexity of conventional Transformer architectures and their considerable training data demands, Bao et al. \cite{bao2025lightweight} introduced a streamlined RUL prediction framework that combines signal decomposition with a lightweight and term-arbitrary memory network (LTM-Net).

Signal decomposition brings new challenges. Decomposition algorithms involve parameters which need to be optimized, e.g., with computationally expensive meta-heuristic algorithms \cite{9758685,CHEN2024113388}. Besides, generic ML models are not customized to frequency-specific characteristics and the prognostics performance is often suboptimal. To address the challenges, several hybrid models have been proposed to process the decomposed signals or incorporate physics-based or empirical models. Liang et al. \cite{liang2024hybrid} developed a method to combine double exponential model with GRU-CNN networks, and the outputs are processed by a Bayesian neural network. Xie et al. \cite{xie2020prognostic} proposed a method based on particle filter (PF) and LSTM. Zhang et al. \cite{zhang2023data} introduced an integrated BiGRU and temporal self-attention mechanism (TSAM) with PF. Lu et al. \cite{lu2025remaining} developed empirical model-based approaches with improved PF algorithms, and Chen et al. \cite{chen2024hybrid} combined Arrhenius principles with lightweight Transformers. For many of these methods, the integration is static and real-time adaptation is unavailable \cite{liang2024hybrid,zhang2023data}. As such, decomposed signals are inadequately utilized and the models are not continuously refined for practical battery prognostics.


To address these limitations, we propose \textsc{Karma}, a \underline{k}nowledge-\underline{a}wa\underline{r}e \underline{m}odel with frequency \underline{a}daptive learning for battery health prognostics. First, we develop a frequency-adaptive dual-stream architecture for battery capacity prediction. Raw battery signals are decomposed into different frequency bands and we group them into low- and high-frequency signals. CNN-LSTM is employed for low-frequency signals for its strength in capturing long-term degradation trends. BiGRU is used for high-frequency signals which model rapid battery dynamics and noises. The two streams are integrated with the attention mechanism and future battery capacity can be estimated. The dual-stream design overcomes a challenging if not impossible task of customizing a single model for different frequency bands. Note that CNN-LSTM and BiGRU can be upgraded or replaced with more suitable ML algorithms. We aim to demonstrate the capability of our frequency adaptive design, and our choices are not based on exhaustive search of different algorithms. While data-driven approaches capture rich temporal and frequency patterns, they often lack physical grounding, which leads to unreliable long-term forecasts. In \textsc{Karma}, we embed empirical knowledge of battery degradation through a double exponential function, which characterizes typical nonlinear fading patterns of battery degradation. The function involves several parameters and we employ PF to optimize the parameters, by aligning them with both observed/historical battery data and data-driven model predictions. The final SoH and RUL predictions are based on the knowledge instead of the data-driven dual-stream model. Such as knowledge-regulated design ensures forecasts remain physically meaningful with accurate and robust predictions for different datasets and operating conditions. Furthermore, PF in \textsc{Karma} realizes uncertainty quantification which is highly important for industry applications. Specifically, we have the following main contributions in this paper.

\begin{itemize}
    \item We propose \textsc{Karma}, a novel hybrid framework that integrates data-driven learning with knowledge-based modeling for battery capacity estimation and RUL prediction to improve accuracy and robustness across diverse degradation patterns and operating conditions. 
    \item We propose a frequency adaptive learning model with a dual-stream architecture for battery signals in low and high frequency bands, corresponding to long-term degradation trends and short-term dynamics, respectively.
    \item We conduct experimental study for \textsc{Karma} in different settings. \textsc{Karma} achieves on average 50.6\% and 32.6\% error reductions for two mainstream datasets, respectively, compared to several latest comparison algorithms. 
\end{itemize}

The remainder of this paper is structured as follows. Section \ref{sec:method} presents the methodology of \textsc{Karma}. Section \ref{sec:exp} provides experimental results and analysis. Finally, Section \ref{sec:conclusion} concludes the paper and suggests future works.

\section{\textsc{Karma}: Methodology}
\label{sec:method}
\begin{figure}[]
    \centering
    \includegraphics[width=0.98\linewidth]{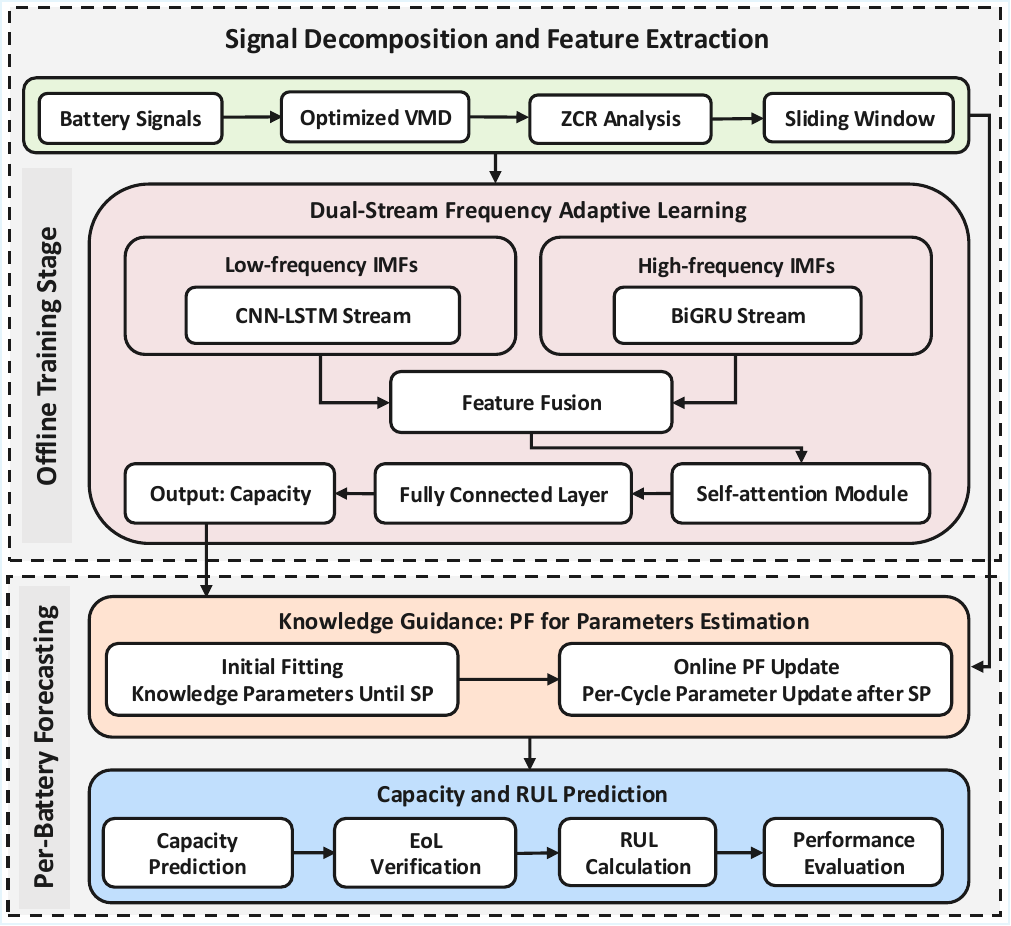}
    \caption{An illustration of \textsc{Karma}'s system architecture. It includes signal decomposition module, a dual-stream frequency adaptive model and a knowledge-based degradation model, which outputs the battery capacity estimation and RUL prediction.}
    \label{fig:sys}
\end{figure}

In this section, we present the detailed methodology of \textsc{Karma}. First, we illustrate the overall workflow of \textsc{Karma} in Fig. \ref{fig:sys}. As shown in the figure, \textsc{Karma} is a hybrid framework for battery health prognostics that combines frequency-aware deep learning with knowledge-guided degradation modelling. The framework consists of a training stage and a per-battery forecasting stage. In the training stage, we begin with a training dataset with raw battery signals such as sensor readings. We apply signal decomposition to extract multiple frequency bands, to capture long-term degradation trends and short-term fluctuations. Each band is represented as an intrinsic mode function (IMF), and we develop a dual-stream frequency-adaptive model to process them. One stream is based on CNN-LSTM for learning long-term degradation trends, while the other leverages BiGRU to capture short-term fluctuations and residual dynamics. This design enables the model to be both trend-aware and responsive and accordingly enhances predictive performance at the system level. The offline trained model is directly applied to any new target battery in the next stage as detailed below.

In the per-battery forecasting stage, \textsc{Karma} enables dynamic and cycle-by-cycle battery health estimation. \textsc{Karma} leverages empirical battery knowledge and models the degradation trajectory as a double exponential function. Given a target battery, \textsc{Karma} first uses the observed capacity values of the battery up to the current cycle to estimate the parameters of the function using a PF algorithm. Beyond the current cycle, the pre-trained dual-stream model is used to predict future capacity values, which are combined with the real observations before the cycle to recursively update the function parameters. This process continues cycle by cycle, updating the degradation trajectory and health estimate at each step, until the projected function reaches the battery’s EoL. The RUL can then be calculated, e.g., the number of cycles between the current and EoL cycles. In summary, \textsc{Karma} integrates frequency-adaptive learning with knowledge-aware modeling to enable accurate battery prognostics. The following parts present the technical details of \textsc{Karma}.

\subsection{Signal Decomposition into Frequency-Aware IMFs}
The first task of \textsc{Karma} is to decompose raw battery signals into different frequency bands. Among various signal decomposition techniques, we adopt VMD, one of the most widely used algorithms in the recent literature in this domain. The technical details of VMD is available in our previous work \cite{pamshetti2025optimal} and we only present it briefly in this paper. Battery signals are time-series-based and VMD is applied to decompose each time-series into a set of IMFs, each is characterized by a frequency band. The two optimization objectives of VMD are, minimizing the overlap between different IMFs, and reconstructing the raw signal accurately with the derived IMFs. Let $f(\cdot)$ be the raw signal. The optimization problem is,
\begin{equation}
\begin{aligned}
\min_{u_k,\omega_k} & \left\{ \sum\nolimits_{k=1}^{k^{\max}} \left\| \frac{\partial}{\partial t} \left[ \Big(\delta(t) + \frac{j}{\pi t}\Big) * u_k(t) \right] e^{-j \omega_k t} \right\|_2^2 \right\}, \\
& \quad\quad \text{subject to } \sum\nolimits_{k=1}^{k^{\max}} u_k(t) = f(t),
\end{aligned}
\label{eq:vmd-prob}
\end{equation}
where $k^{\max}$ is the number of resulting IMFs and $u_k$ is the $k$-th IMF. Let $\omega_k$ be the central frequency of $u_k$. VMD aims to minimize the total bandwidth of all IMFs for minimizing overlap, where the bandwidth of each $u_k$ is derived using the Hilbert transform kernel $\delta(t) + \frac{j}{\pi t}$. $\delta(t)$ represents the original signal and $\frac{j}{\pi t}$ is the Hilbert transform of the signal. $u_k$ is convoluted with the kernel, where $e^{-j \omega_k t}$ is used for frequency shift with $\omega_k$ to shift the signal to a near-zero frequency band. Then, the squared L2-norm $\left\|\cdot \right\|_2^2$ measures the signal magnitude. The constraint is introduced to make sure that the raw signal can be reconstructed by all IMFs. The problem in Eq. (\ref{eq:vmd-prob}) can be solved by an augmented Lagrangian as,
\begin{equation}
\begin{aligned}
\mathcal{L}(u_k, \omega_k, \lambda) =& \alpha \sum_{k=1}^{k^{\max}} \left\| b_k \right\|_2^2 + \left\| f(t) - \sum_{k=1}^{k^{\max}} u_k(t) \right\|_2^2 \\
&+ \langle \lambda(t), f(t) - \sum_{k=1}^{k^{\max}} u_k(t) \rangle,
\end{aligned}
\label{eq:vmd-lagrangian}
\end{equation}
where the first two terms capture overlap minimization and constraint, respectively. The two terms or objectives can be contradictory, so we balance them with a parameter $\alpha$. The third term with the Lagrange multiplier enforces the constraint and we use $\lambda$ to control the impact of such a regulation term. 

Based on Eqs. (\ref{eq:vmd-prob}) and (\ref{eq:vmd-lagrangian}), two VMD parameters are the number of IMFs $k^{\max}$ and $\alpha$. The parameters should be carefully chosen, as overly large or small values may degrade decomposition quality. Here, we follow our work in \cite{pamshetti2025optimal} and apply particle swarm optimization (PSO) to search for the optimal parameters. We would like to mention our observation that the optimal parameters can be derived by several meta-heuristics, including PSO. Replacing PSO with another meta-heuristic does not yield significant performance improvements or research novelty in the context of battery health prognostics.

\subsection{Frequency-Aware and Temporal Feature Extraction}
Following signal decomposition, we have $k^{\max}$ IMFs characterized by different frequency bands. To distinguish meaningful degradation trends from high-frequency noises, we perform frequency-based classification using zero-crossing rate (ZCR) analysis \cite{li2023hybrid} which quantifies signal oscillatory. The ZCR of each IMF is calculated as $\texttt{ZCR}(u_k)=n_{\text{zero}}/ n$, where the numerator is the number of sign changes and the denominator is the signal length. We follow an empirical setting where IMFs with ZCR values exceeding 0.01 are classified as high-frequency components, typically associated with rapid fluctuations or measurement noise. In contrast, IMFs with lower rates are considered as low-frequency components, which capture gradual degradation patterns and long-term battery aging. The IMF classification enables \textsc{Karma} to process different frequency behaviors with specialized models.

After classification, we transform the IMFs into structured temporal sequences using a sliding window approach for time-series based battery health forecasting. The sequences allow an ML model to learn temporal dependencies across cycles, and this is important for sequential battery degradation prediction. We then scale each windowed sequence using \texttt{min-max} normalization to ensure consistent magnitude across all input channels. As such we can prevent dominant IMFs from skewing the learning process. Overall, through structured and normalized representation, \textsc{Karma} generates frequency-aware and temporal input features that preserve both short-term dynamics and long-term trends for the downstream dual-stream ML architecture. 

\begin{figure}[]
\centering
\includegraphics[width=0.98\linewidth]{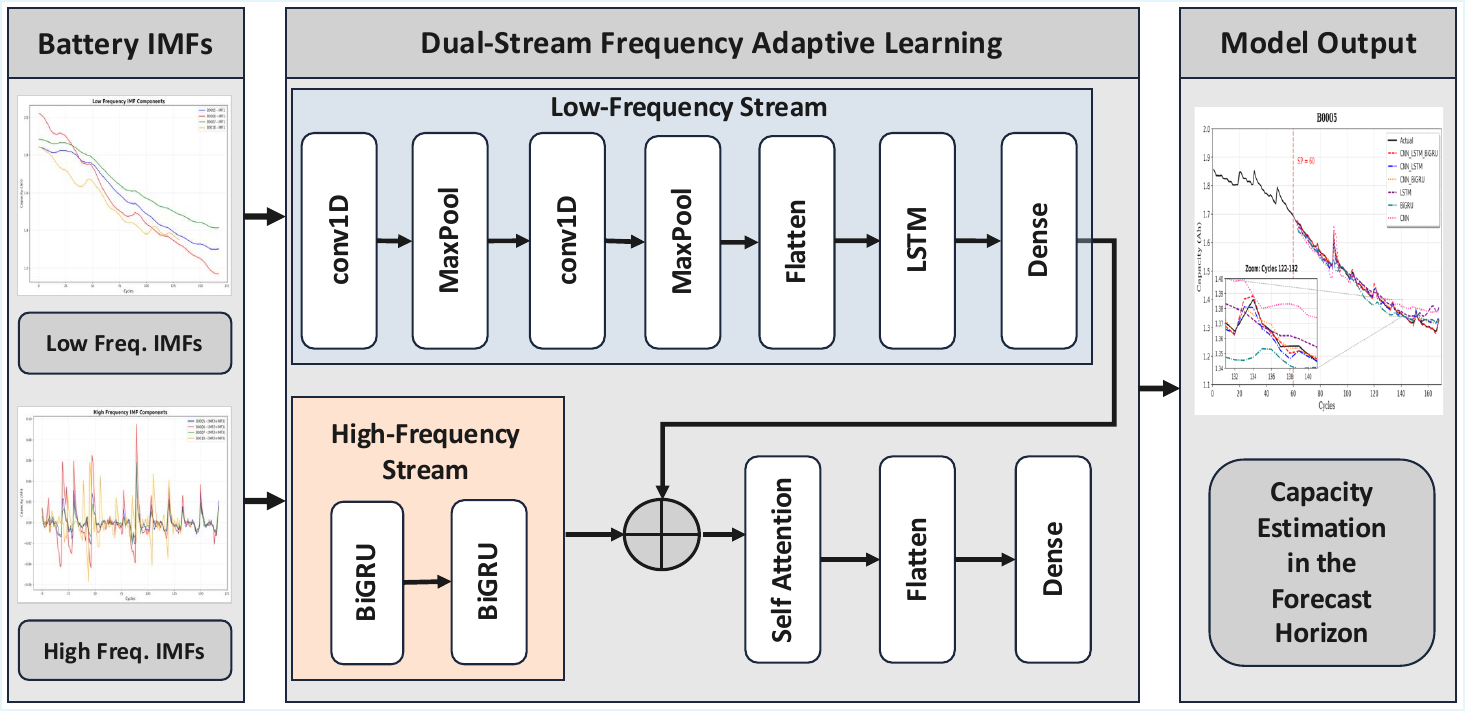}
\caption{An illustration of the proposed dual-stream frequency adaptive learning model for battery capacity estimation in a given forecast horizon. The two parallel streams are based on CNN-LSTM and BiGRU for low- and high-frequency IMFs, respectively. The streams are fused into one before final processing including self-attention and model output.}
\label{fig:dual-stream-ml}
\end{figure}

\subsection{A Dual-Stream Frequency-Aware Model}
Given the low- and high-frequency IMFs obtained from the previous step, we construct two separate ML streams for specialized processing as shown in Fig. \ref{fig:dual-stream-ml}. Let $X^L$ and $X^H$ be the sets of low- and high-frequency IMFs. One stream applies a CNN-LSTM pipeline to $X^L$ as,
\begin{equation}
    \mathbf{f}^{L} = \texttt{LSTM}\big(\texttt{CNN}(X^{L})\big),
\end{equation}
where two convolutional layers with max-pooling extract local temporal patterns over extended time spans, and one LSTM layer models long-term dependencies associated with gradual capacity fade. The high-frequency $X^H$ is processed in the other stream based on BiGRU as,
\begin{equation}
    \mathbf{f}^{H} = \texttt{BiGRU}(X^{H}),
\end{equation}
which captures short-term variations and transient dynamics by analyzing the data in both forward and backward directions. High-frequency $X^H$ is dominated by rapid fluctuations, where long-term memory, e.g., in LSTM, is less essential compared to processing $X^L$. As such, BiGRU, which has a lightweight architecture, is suitable and chosen for this stream.

The outputs of the dual-stream are $\mathbf{f}^L$ and $\mathbf{f}^H$ and we perform feature fusion to integrate them. We first concatenate the outputs as $\mathbf{f} = \texttt{concat}(\mathbf{f}^L, \mathbf{f}^H)$. Then we pass the fused feature $\mathbf{f}$ into an attention layer to adaptively weight the features as,
\begin{equation}
    \texttt{Attention}(Q,K,V) = \texttt{softmax}\left( QK^\top / \sqrt{d_k} \right)V,
\end{equation}
where $Q$, $K$ and $V$ are learned projections based on the fused feature $\mathbf{f}$ and corresponding weight matrices and $d_k$ is the dimension of the keys. This attention enables the model to focus on the most relevant temporal patterns for the current battery health prediction. Finally, the attention-weighted features are pass through fully connected layers to generate $\hat{y}$ as the capacity estimate. Overall, our dual-stream architecture combines frequency-specific processing with adaptive fusion and can capture the multi-scale nature of battery degradation with improved prediction accuracy.

\subsection{Knowledge-based Degradation Model}
The dual-stream model presented above is purely data-driven. Same as most data-driven models, its predictions may drift over long forecasting horizons due to various reasons such as noise accumulation. As such, \textsc{Karma} employs domain knowledge to guide the forecasting process. Specifically, we follow the empirical knowledge and adopt a double exponential capacity decay function \cite{zhang2023data}. The function reflects two widely reported empirical patterns, including fast degradation process such as formation loss and early performance drop, as well as slow degradation process such as gradual material aging. The function is determined by physically meaningful parameters, including amplitudes and decay rates, that are interpretable and can be customized for different batteries. To search for the optimal parameters for each battery, we utilize real measurements at the beginning stage of a battery and then apply the developed dual-stream prediction model to continuously adapts the degradation trajectory, i.e., the function, to match the measurements and predictions. We present the technical details of the function and parameter estimation algorithm as follows.

\subsubsection{Double Exponential Function}
The function assumes that battery capacity $C(k)$ at cycle $k$ can be expressed as the sum of two exponential terms as,
\begin{equation}
    C(k) = a \cdot \exp(b \cdot k) + c \cdot \exp(d \cdot k),
\end{equation}
where each term represents a distinct degradation process. The first term is parameterized by $a$ and $b$ to capture early-life capacity loss. The second term parameterized by $c$ and $d$ models slow and progressive degradation driven by long-term mechanisms such as micro-structural damage accumulation and electrolyte decomposition. Among the four parameters, $a$ and $c$ are positive amplitudes reflecting the magnitude of each process and $b$ and $d$ are negative decay rates. For example, $|b| >|d|$ implies fast initial decline and slow long-term aging. Note that the function serves as a constraint on the dual-stream data-driven model and ensures that long-horizon forecasting remains knowledge-guided. 

\subsubsection{Parameter Estimation with PF}
Given the function, our objective now is to estimate and update the four parameters  for each specific battery in a sequential manner. We adopt the classic PF algorithm for its capability of handling nonlinear and non-Gaussian systems while maintaining a probabilistic representation of uncertainty, which is important for battery analytics, in parameter estimates. PF operates on two core mathematical relationships as below. 

\paragraph{State Evolution Model} 
We treat the parameter vector $\boldsymbol{\theta}_k = (a_k, b_k, c_k, d_k)$ as the state of the system at cycle $k$. While the parameters are expected to evolve slowly over time, small variations exist because of operating condition changes and modeling inaccuracies. Thus we have,
\begin{equation}
    \boldsymbol{\theta}_k \sim f^\text{S}(\boldsymbol{\theta}_{k-1}) + \mathcal{N}(0, \mathbf{Q}^{\text{PF}}),
    \label{eq:state-evo}
\end{equation}
where the parameter vector $\boldsymbol{\theta}_k$ is approximated as the previous estimate passed through the state transition function $f^\text{S}(\cdot)$, perturbed by Gaussian process noise with covariance $\mathbf{Q}^{\text{PF}}$.

\paragraph{Measurement Model}
For each particle with parameters $\boldsymbol{\theta}_k$ at cycle $k$, the knowledge-based capacity estimation is $\hat{C}(k)$. To align the knowledge-based model with a battery's degradation trajectory, we involve a measurement model. Our aim is to compare $\hat{C}(k)$ with the capacity observation $\hat{y}_k$ by the dual-stream data-driven model. The difference between the two estimations is the measurement error $v_k$ as,
\begin{equation}
    v_k = \hat{y}_k - \hat{C}(k),
    \label{eq:mesurement-error}
\end{equation}
where the error is used to adjust the particle's parameters. Among many particles in the PF algorithm, the ones that produce small errors are given high weights, while the rest with large errors are relatively down-weighted. Over optimization iterations, particles or parameter sets that do not match the dual-stream estimations are filtered out, and only the ones with consistent alignment between knowledge-based and data-driven estimations remain. 


\subsubsection{PF Algorithm Workflow}
Here, we present the PF algorithm workflow specialized for battery degradation modelling with the following key steps.

\paragraph{PF Parametric Initialization}
We denote the current cycle as the prediction start point (SP), until when, the real battery capacity measurements are known and corresponding IMFs can be calculated accordingly. We first fit the knowledge-based model to the real measurements until cycle SP to obtain the initialized parameters $\boldsymbol{\theta}_{SP}$. Let $y_k$ be the real capacity at cycle $k$ before SP, we estimate $\boldsymbol{\theta}_{SP}$ by solving a constrained least-squares problem over a fitting window from cycle $k^{\text{ini}}$ to cycle SP, where $k^{\text{ini}}>0$ can be customized to exclude potential highly noisy early measurements. We have,
\begin{equation}
    \hat{\boldsymbol{\theta}}_{SP} \;
    =\; \arg\min_{\boldsymbol{\theta}\in\Theta} \sum\nolimits_{k=k^{\text{ini}}}^{\text{SP}} \Big(g(\boldsymbol{\theta},k) - y_k\Big)^2,
\end{equation}
where $g(\boldsymbol{\theta},k)=a \cdot \exp(b \cdot k) + c \cdot \exp(d \cdot k)$ with $\boldsymbol{\theta}=(a,b,c,d)$. $\Theta$ is the feasible set of the parameters to enforce physically meaningful constraints, e.g., $a,c>0$, $b,d<0$, and $a+c \approx y_0$. Through the optimization, we aim to find an optimized $\hat{\boldsymbol{\theta}}_{SP}$ with the expectation that a good initialization contributes to accelerated convergence of the battery health model. Besides $\hat{\boldsymbol{\theta}}_{SP}$, we produce $\boldsymbol{\Sigma}_{SP}$ to quantify the parameter uncertainty, which is important for initialing particles as below.

\paragraph{Particle Initialization}
With $\hat{\boldsymbol{\theta}}_{SP}$ and $\boldsymbol{\Sigma}_{SP}$, we initialize a set of $n$ particles to represent the posterior distribution of the parameters. Each particle $i\in [1,n]$ is sampled from a Gaussian around the fitted estimate as, 
\begin{equation}
\boldsymbol{\theta}_{SP}^{i} = \hat{\boldsymbol{\theta}}_{SP} + \boldsymbol{\epsilon}_i,
\end{equation}
where $\boldsymbol{\epsilon}_i \sim \mathcal{N}(\mathbf{0}, \boldsymbol{\Sigma}_{SP})$ represents initialization noise with covariance matrix $\boldsymbol{\Sigma}_{SP}$. Such a sampling scheme offers diversity into the particles so the realistic parameter uncertainty can be explored. In practice, the matrix should be configured to balance exploration and stability, e.g., a small covariance reduces adaptability and a large covariance distributes particles on implausible parameter regions. Similar to the above step, we ensure physical plausibility and project each particle as,
\begin{equation}
\boldsymbol{\theta}_{SP}^{i} \leftarrow \Pi_{\Theta}(\boldsymbol{\theta}_{SP}^{i}),
\end{equation}
where $\Pi_{\Theta}(\cdot)$ is the projection operator for feasible parameter set $\Theta$. Initial particle weights are set uniformly and the weight of particle $i$ is $w^{i} = 1/n$.

\paragraph{Propagation}
At each forecasting cycle $k > \text{SP}$, the parameters of each particle are propagated forward according to the state evolution model in Eq. (\ref{eq:state-evo}). Variations are allowed to reflect operating condition changes and the process noise is drawn from $\mathcal{N}(\mathbf{0}, \mathbf{Q}^{\text{PF}})$. The benefit of the noises is that most particles can explore the space around their previous states, so that the risks of premature convergence to suboptimal can be minimized. Afterwards, particle parameters are projected back to the feasible set $\Theta$.

\paragraph{Weight Update}
After propagation, each particle's parameters $\boldsymbol{\theta}_k^{i}$ are evaluated against the data-driven model. Based on Eq. (\ref{eq:mesurement-error}), we calculate the measurement error $v_k^{i}$ for particle $i$ at cycle $k$. The particle's weight is updated as,
\begin{equation}
    w_k^{i} = w_{k-1}^{i} \cdot p\big(\hat{y}_k \mid \boldsymbol{\theta}_k^{i}\big),
\end{equation}
where the likelihood function $p(\cdot)$ is modeled as a Gaussian,
\begin{equation}
p\big(\hat{y}_k \mid \boldsymbol{\theta}_k^{i}\big) \propto \exp\bigg( -\frac{\big(v_k^{i}\big)^2}{2\sigma_{\text{err}}^2} \bigg),
\end{equation}
where $\sigma_{\text{err}}^2$ is the measurement error variance. Finally, we normalize the weights of all particles and the weight of particle $i$ is $w_k^{i}\gets w_k^{i}/\sum_{i=1}^n w_k^{i}$. Intuitively, we prefer particles with small errors and assign high weights to them, and we expect the probability mass align with both knowledge-based and data-driven models.

\paragraph{Adaptive Resampling}
Over time, a few particles may carry almost all the weight, while the rest contribute very little. This situation, called degeneracy, implies that most particles are no longer useful for searching optimal parameters. We aim to keep the particles diverse and we monitor the effective sample size (ESS) after weight normalization as,
\begin{equation}
    \texttt{ESS}_k = 1 / \sum\nolimits_{i=1}^n \left(w_k^{i}\right)^2,
\end{equation}
where a high (low) ESS indicates that weights are spread out evenly (concentrated). If ESS drops below a pre-defined threshold $\tau_\text{ESS}$, we perform multinomial resampling by duplicating (discarding) high (low) weights particles. Specifically, this adaptive resampling may not be triggered every cycle, and we typically set $\tau_\text{ESS}=n/2$ with resampling being triggered when $\texttt{ESS}_k < \tau_\text{ESS}$. Furthermore, we initialize the weight of every particle $i$ to $w_k^{i} \gets 1/n$ for $n$ particles. As a result, we have a new set of particle with improved coverage of the probable regions of the state space.

\paragraph{State Estimation and Capacity Prediction}
After the above steps, the current state is estimated by combining the information from all particles. The final parameter estimates are computed as,
\begin{equation}
\hat{\boldsymbol{\theta}}_k = \sum\nolimits_{i=1}^{n} w_k^{i} \boldsymbol{\theta}_k^{i},
\end{equation}
where $\hat{\boldsymbol{\theta}}_k$ is then used as the final parameters to estimate the capacity $\hat{C}(k)$. Uncertainty at cycle $k$ can be computed based on the per-particle capacities and weights $\{C^i(k), w_k^{i}\}_{i=1}^n$. 

At each cycle $k$, we check if the estimated capacity $\hat{C}_k$ has dropped below the EoL threshold, i.e., $C^{\text{Eol}}$ where one setting is 70\% of the battery's original maximum capacity. If true, the RUL is computed as,
\begin{equation}
\texttt{RUL}_{\text{SP}} = k - \text{SP},
\end{equation}
and the PF process terminates. Otherwise, we generate the predicted capacity with the data-driven model for the next cycle $k+1$, and update the particle parameters based on the above steps. Note that the particles are not re-initialized and the weights in cycle $k$ are inherited in cycle $k+1$. The updated parameters are used to compute the capacity $\hat{C}(k+1)$, after which the EoL check is repeated. This iterative process continues until the EoL condition is met.

\section{Experimental Study}
\label{sec:exp}
We conduct a comprehensive experimental study to evaluate the performance of \textsc{Karma} and present the results and discussions in this section.

\subsection{Experiment Setup}
We first present experiment setup, including datasets and performance evaluation metrics.

\subsubsection{Data}
This study employs two established Li-ion battery aging datasets, produced by NASA's Prognostics Center of Excellence \cite{saha2007battery} and the University of Maryland's Center for Advanced Life Cycle Engineering (CALCE) \cite{calce2017lithium}. We denote the first one as the NASA dataset, which comprises comprehensive charge/discharge aging experiments conducted on 18650-type Li-ion batteries with 2 Ah nominal capacity. We consider four batteries, including B0005, B0006, B0007 and B0018. The battery testing protocol involves cyclical charging, discharging and electrochemical impedance spectroscopy measurements at 24$^{\circ}$C ambient temperature. The charging procedure includes a constant current protocol (1.5 A to 4.2 V) followed by constant voltage charging (until current $\leq$ 20 mA), while discharging has constant current until terminal voltages reach 2.7~V, 2.5~V, 2.2~V and 2.5~V for the four batteries. The dataset defines EoL as 70\% capacity retention relative to the rated capacity, corresponding to 1.40~Ah threshold. One exception is battery B0007, with more than 1.40 Ah capacity remaining after the completion of the battery testing. We customize the EoL criterion to 72\%, i.e., 1.44 Ah, to facilitate performance analysis. The CALCE dataset features graphite anode/LiCoO$_2$ cathode Li-ion batteries with 1.1~Ah nominal capacity. Four batteries, including CS35, CS36, CS37 and CS38, operating at 1C discharge rate are selected for this study. The battery testing of this dataset is completed with an Arbin BT2000 system at room temperature with 0.5C charging to 4.2~V, followed by constant voltage charging until current dropped below 0.05~A. Discharging applies 1C rate until 2.7~V. Same as the NASA dataset, EoL threshold is 70\% of capacity retention, i.e., 0.77~Ah.

\begin{table*}[]
\centering
\renewcommand{\arraystretch}{1.3}
\caption{Performance comparison of different battery health models on both battery dataset in terms of MAE ($\times 10^{-2}$), RMSE ($\times 10^{-2}$), MAPE (\%) and the average MAPE (\%) for all batteries, where low values indicate good performance. Results show that our \textsc{Karma} outperforms comparison algorithms and achieves the lowest errors across all batteries.}
\begin{tabular}{c|ccc|ccc|ccc|ccc|c}
\hline\hline
\multirow{3}{*}{\shortstack{Battery Model\\1-Cycle Capacity}} & \multicolumn{13}{c}{NASA Dataset}\\ \cline{2-14}
 & \multicolumn{3}{c|}{B0005} 
 & \multicolumn{3}{c|}{B0006} 
 & \multicolumn{3}{c|}{B0007} 
 & \multicolumn{3}{c|}{B0018} 
 & \multirow{2}{*}{\shortstack{Average\\MAPE}} \\ 
\cline{2-13}
 & MAE & RMSE & MAPE & MAE & RMSE & MAPE & MAE & RMSE & MAPE & MAE & RMSE & MAPE &  \\ 
\hline\hline
\textsc{Karma} (ours) & \textbf{0.323} & \textbf{0.450} & \textbf{0.227} & \textbf{0.534} & \textbf{0.666} & \textbf{0.382} & \textbf{0.236} & \textbf{0.352} & \textbf{0.152} & \textbf{0.536} & \textbf{0.742} & \textbf{0.374} & \textbf{0.284} \\ 
\hline
CNN-LSTM   & 0.890 & 1.167 & 0.623 & 1.198 & 1.606 & 0.884 & 0.681 & 0.913 & 0.442 & 1.052 & 1.392 & 0.739 & 0.672 \\ 
\hline
CNN-BiGRU  & 0.969 & 1.216 & 0.680 & 1.391 & 1.797 & 1.024 & 0.843 & 1.214 & 0.565 & 1.303 & 1.731 & 0.917 & 0.796 \\ 
\hline
LSTM        & 1.804 & 2.531 & 1.295 & 1.940 & 3.696 & 1.511 & 2.384 & 2.923 & 1.668 & 2.384 & 2.923 & 1.668 & 1.536 \\ 
\hline
BiGRU       & 1.918 & 2.442 & 1.362 & 3.249 & 3.870 & 2.398 & 1.405 & 1.873 & 0.906 & 1.918 & 2.442 & 1.362 & 1.507 \\ 
\hline
CNN         & 2.697 & 3.229 & 1.911 & 2.990 & 3.744 & 2.219 & 2.878 & 4.131 & 2.256 & 2.107 & 2.553 & 1.503 & 1.972 \\ 
\hline\hline
\multirow{3}{*}{\shortstack{Battery Model\\1-Cycle Capacity}} & \multicolumn{13}{c}{CALCE Dataset}\\ \cline{2-14}
 & \multicolumn{3}{c|}{B0005} 
 & \multicolumn{3}{c|}{B0006} 
 & \multicolumn{3}{c|}{B0007} 
 & \multicolumn{3}{c|}{B0018} 
 & \multirow{2}{*}{\shortstack{Average\\MAPE}} \\ 
\cline{2-13}
 & MAE & RMSE & MAPE & MAE & RMSE & MAPE & MAE & RMSE & MAPE & MAE & RMSE & MAPE &  \\ 
\hline\hline
\textsc{Karma} (Ours) & \textbf{0.415} & \textbf{0.656} & \textbf{0.613} & \textbf{0.569} & \textbf{0.867} & \textbf{1.073} & \textbf{0.465} & \textbf{0.694} & \textbf{0.724} & \textbf{0.439} & \textbf{0.653} & \textbf{0.615} & \textbf{0.756} \\ \hline
CNN-LSTM       & 1.210 & 1.714 & 1.829 & 1.221 & 2.127 & 3.727 & 0.918 & 1.247 & 1.668 & 0.745 & 0.954 & 1.024 & 2.062 \\ \hline
CNN-BiGRU      & 1.159 & 1.649 & 1.742 & 1.437 & 1.897 & 3.187 & 1.038 & 1.490 & 2.025 & 1.075 & 1.409 & 1.655 & 2.152 \\ \hline
LSTM            & 1.663 & 4.219 & 3.150 & 2.345 & 6.007 & 7.739 & 1.046 & 1.385 & 1.612 & 2.241 & 2.519 & 2.978 & 3.870 \\ \hline
BiGRU           & 2.207 & 3.351 & 4.099 & 2.946 & 3.966 & 6.606 & 1.328 & 1.937 & 2.411 & 2.491 & 3.048 & 3.662 & 4.194 \\ \hline
CNN             & 1.448 & 1.976 & 2.331 & 1.608 & 2.321 & 3.981 & 1.451 & 1.958 & 2.541 & 1.318 & 1.644 & 1.877 & 2.682 \\ \hline\hline
\end{tabular}
\label{tab:comp-soh}
\end{table*}

\begin{figure*}[]
    \centering
    \def \tmph{1.45in}
    \subfigure[NASA B0005]{\includegraphics[height=\tmph]{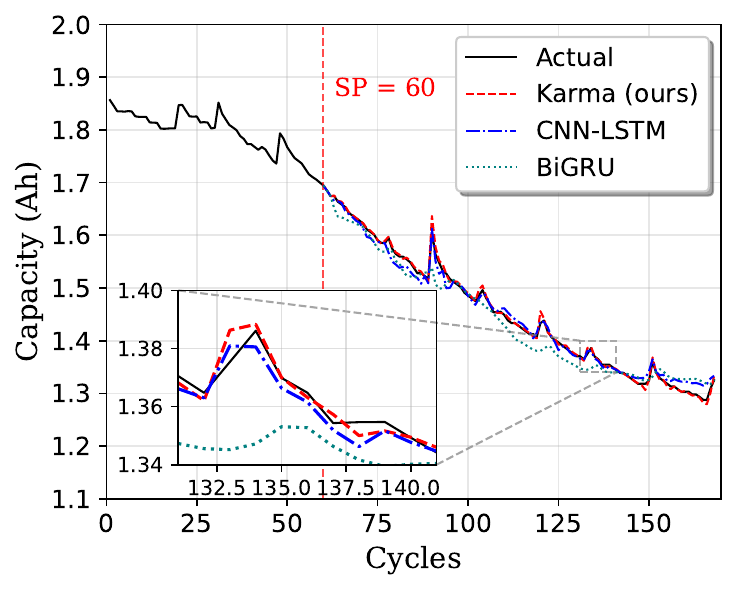}\label{fig:comp-soh-nasa-curve}}
    \hspace{-0.2cm}
    \subfigure[NASA B0005]{\includegraphics[height=\tmph]{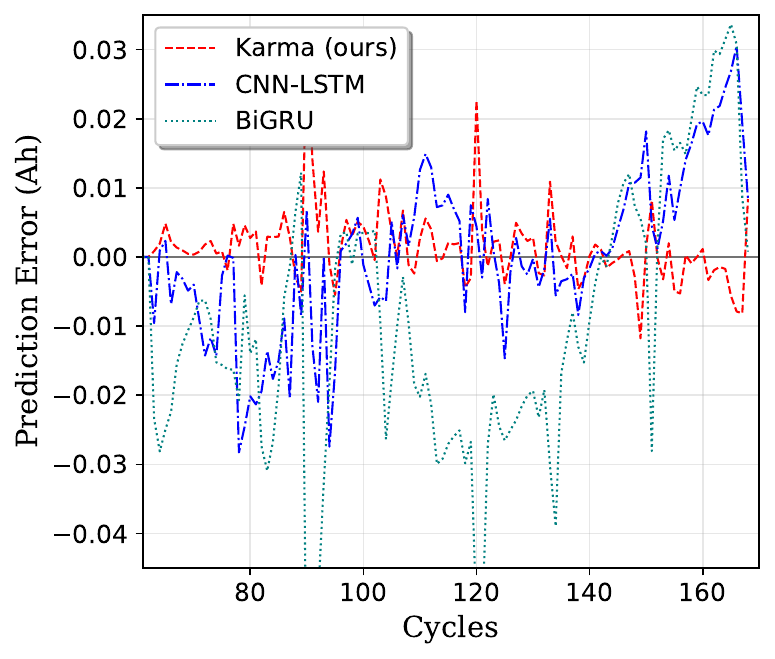}\label{fig:comp-soh-nasa-err}}
    \hspace{-0.2cm}
    \subfigure[CALCE C35]{\includegraphics[height=\tmph]{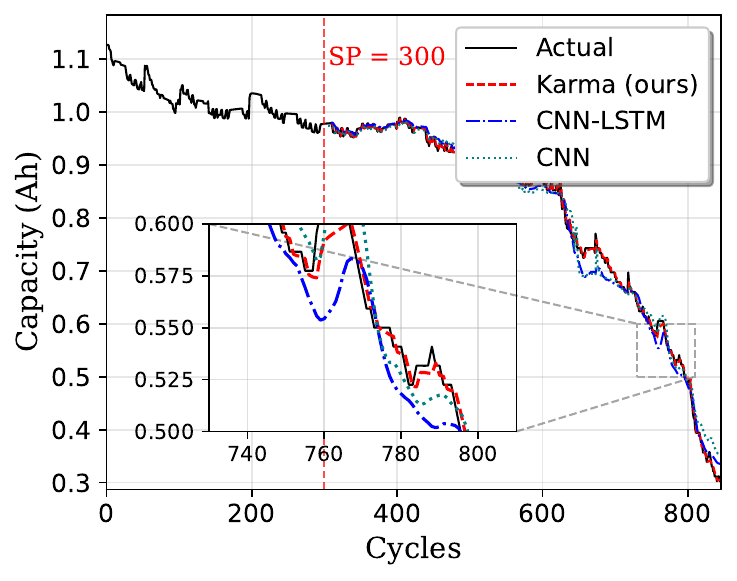}\label{fig:comp-soh-cs-curve}}
    \hspace{-0.2cm}
    \subfigure[CALCE C35]{\includegraphics[height=\tmph]{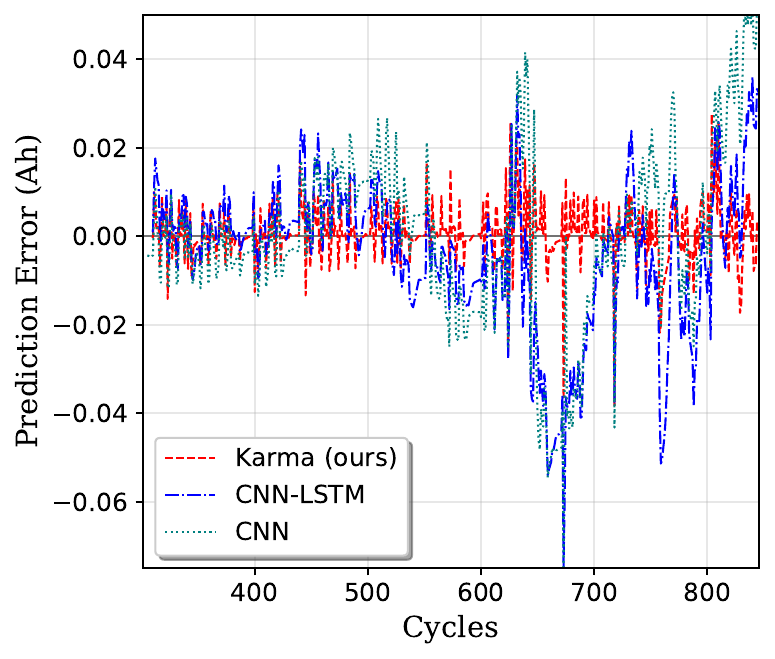}\label{fig:comp-soh-cs-err}}
    \caption{The degradation trajectories of the first NASA battery and the first CALCE battery in Figs. \ref{fig:comp-soh-nasa-curve} and \ref{fig:comp-soh-cs-curve}, respectively, with capacity estimations from \textsc{Karma}, CNN-LSTM and BiGRU. The corresponding prediction errors are visualized in Figs. \ref{fig:comp-soh-nasa-err} and \ref{fig:comp-soh-cs-err}, respectively. \textsc{Karma} aligns with the ground-truth the best with minimized prediction errors.}
    \label{fig:comp-soh-cs}
\end{figure*}

\subsubsection{Performance Metrics} 
We evaluate the performance of different algorithms using comprehensive performance metrics, including mean absolute error (MAE), root mean square error (RMSE) and mean absolute percentage error (MAPE). Given a list of predicted values $\hat{y}_1, \ldots, \hat{y}_n$ and corresponding ground-truth $y_1, \ldots, y_n$, MAE is calculated as,
\begin{equation}
\texttt{MAE} = \frac{1}{n} \sum\nolimits_{i=1}^{n} |y_i - \hat{y}_i|,
\end{equation}
where $|y_i - \hat{y}_i|$ is the absolute error between the predicted value and ground-truth. RMSE is calculated as,
\begin{equation}
\texttt{RMSE} = \sqrt{\frac{1}{n} \sum\nolimits_{i=1}^{n} (y_i - \hat{y}_i)^2},
\end{equation}
which, compared to MAE, is affected by outliers more significantly. We also quantify the percentage difference with MAPE, given by,
\begin{equation}
\texttt{MAPE} = \frac{1}{n} \sum\nolimits_{i=1}^{n} \left|(y_i - \hat{y}_i) / y_i\right| \times 100\%.
\end{equation}

\subsection{Comparison Study}
We first present our comparison study where \textsc{Karma} is compared against several comparison algorithms. We have two distinct prediction tasks, including 1-cycle ahead capacity prediction and RUL estimation, for NASA and CALCE datasets.

\subsubsection{1-Cycle Capacity Prediction}
For this short-term prediction task, we compare the proposed \textsc{Karma}, which has a hybrid ML architecture, with individual ML models, including LSTM, BiGRU and CNN. Based on existing studies, the three models are good at sequential modelling, extracting convolutional features and capturing temporal dependencies, respectively. Furthermore, we consider hybrid models as well, including CNN-LSTM and CNN-BiGRU, which combine the strengths of different individual ML models.

\paragraph{NASA Data}
For this dataset, we use three out of four batteries to constitute the training data, while the remaining one battery serves as the test data. We refer to existing studies \cite{li2023remaining} and set SP to 60, and the capacity data of the tested battery prior to SP is utilized for training also. The results are shown in Table \ref{tab:comp-soh}, and Figs. \ref{fig:comp-soh-nasa-curve} and \ref{fig:comp-soh-nasa-err} illustrate the degradation trajectory and prediction errors, respectively, of the first NASA battery with capacity estimations from different algorithms.

As we can see, our proposed \textsc{Karma} achieves the best performance across all NASA batteries, with MAE between 0.0024 for B0007 to 0.0054 for B0018. For example, on B0005, \textsc{Karma}'s MAE is 0.0032, an 88\% error reduction compared to CNN's MAE. Against the hybrid models CNN-LSTM and CNN-BiGRU, the error reductions are 63.7\% and 66.7\%, respectively. Overall, the MAPE of \textsc{Karma} is 0.28\% on average, which is much lower than the comparison algorithms. The performance gain arises from \textsc{Karma}'s frequency-adaptive dual-stream architecture, which explicitly separates the processing of low- and high-frequency degradation signals, processed by CNN-LSTM and BiGRU, respectively. \textsc{Karma} also benefits from a knowledge-based model, based on well-understood degradation modes in Li-ion batteries. This hybrid approach avoids the limitations of pure data-driven models, which may overfit noise and lack domain-specific regulation.

Besides, we notice that hybrid models outperform their counterparts with individual models. For example, on B0005, CNN-LSTM achieves an MAE of 0.0089 compared to LSTM's 0.0180 and CNN's 0.0270, with error reductions of 50.6\% and 67.0\%, respectively. CNN-BiGRU similarly improves over BiGRU with 49.5\% error reduction on the same battery. This demonstrates that combining convolutional and recurrent components effectively captures spatial features and temporal dependencies. This brings performance benefits even before introducing \textsc{Karma}'s frequency adaptive enhancements.

\paragraph{CALCE Data}
The CALCE data, similar to the NASA data, comprises Li-ion cells cycled under controlled room-temperature conditions, but with battery chemistry and manufacturing sources different from the NASA data. The differences lead to distinct degradation patterns. The results are shown in Table \ref{tab:comp-soh} and the degradation trends and prediction results for the first CALCE battery are visualized in Figs.~\ref{fig:comp-soh-cs-curve} and \ref{fig:comp-soh-cs-err}, respectively. We observe that \textsc{Karma} outperforms the comparison algorithms despite the differences. For example, on CS36, \textsc{Karma}'s MAE is 0.0057, reducing the errors by 53.4\% and 60.4\% from the MAEs of two hybrid models, CNN-LSTM and CNN-BiGRU which perform better than the three individual ML models, respectively. On CS37, the MAE improves to 0.465, which is 49.3\% and 55.2\% lower than the two models, respectively. These results demonstrate \textsc{Karma}'s competitive performance for a different dataset.

\begin{table*}[t]
\centering
\renewcommand{\arraystretch}{1.3}
\caption{RUL performance comparison of different battery models on both datasets in terms of MAE ($\times 10^{-2}$) and RMSE ($\times 10^{-2}$) for the capacity at EoL cycle with different SPs, where low errors indicate good performance. Results show that \textsc{Karma} outperforms comparison algorithms and achieves the lowest errors for all tested batteries.}
\begin{tabular}{c|ccc|ccc|ccc|ccc|c|c}
\hline\hline
\multirow{3}{*}{\shortstack{Battery Model\\EoL-cycle Capacity}} & \multicolumn{14}{c}{NASA Dataset} \\ \cline{2-15}
& \multicolumn{3}{c|}{B0005} & \multicolumn{3}{c|}{B0006} & \multicolumn{3}{c|}{B0007} & \multicolumn{3}{c|}{B0018} & \multirow{2}{*}{\shortstack{Average\\MAE}} & \multirow{2}{*}{\shortstack{Average\\RMSE}} \\ \cline{2-13}
& SP & MAE & RMSE & SP & MAE & RMSE & SP & MAE & RMSE & SP & MAE & RMSE & & \\ \hline\hline
\multirow{2}{*}{\shortstack{\textsc{Karma}\\(ours)}} 
& 60 & \textbf{0.32} & \textbf{0.42} & 60 & 0.51 & \textbf{0.61} & 50 & \textbf{0.22} & 0.32 & 60 & \textbf{0.53} & \textbf{0.64} & \textbf{0.40} & \textbf{0.50} \\ \cline{2-15}
& 90 & 0.40 & 0.51 & 90 & \textbf{0.44} & \textbf{0.61} & 90 & \textbf{0.22} & \textbf{0.28} & 90 & \textbf{0.53} & \textbf{0.64} & \textbf{0.40} & 0.51 \\ \hline
\textsc{DualStr} & 60 & \textbf{0.32} & 0.45 & 60 & 0.53 & 0.67 & 50 & 0.24 & 0.33 & 60 & 0.54 & 0.74 & 0.41 & 0.55 \\ \hline
Ref. \cite{wang2025exponential} & 61 & 0.67 & 1.25 & 61 & 0.61 & 0.95 & 61 & 0.52 & 0.81 & 61 & 0.54 & 0.75 & 0.59 & 0.94 \\ \hline
Ref. \cite{wang2023remaining} & 67 & 1.01 & 1.45 & 67 & 2.13 & 2.86 & 67 & 0.99 & 1.46 & 67 & 2.03 & 2.52 & 1.54 & 2.07 \\ \hline
Ref. \cite{li2023remaining} & 60 & 0.63 & 1.25 & 90 & 0.72 & 1.15 & 50 & 0.56 & 1.25 & 70 & 1.05 & 2.00 & 0.74 & 1.41 \\ \hline
Ref. \cite{zraibi2021remaining} & 61 & 0.83 & 1.45 & 80 & 0.89 & 1.99 & 54 & 1.20 & 1.72 & 72 & 0.97 & 2.03 & 0.97 & 1.80 \\ \hline
Ref. \cite{li2023hybrid} & 60 & 0.62 & 0.76 & 80 & 0.71 & 0.90 & 50 & 0.64 & 0.77 & 70 & 0.77 & 0.93 & 0.69 & 0.84 \\ \hline\hline
\multirow{3}{*}{\shortstack{Battery Model\\EoL-cycle Capacity}} & \multicolumn{14}{c}{CALCE Dataset} \\ \cline{2-15} 
& \multicolumn{3}{c|}{CS35} & \multicolumn{3}{c|}{CS36} & \multicolumn{3}{c|}{CS37} & \multicolumn{3}{c|}{CS38} & \multirow{2}{*}{\shortstack{Average\\MAE}} & \multirow{2}{*}{\shortstack{Average\\RMSE}} \\ \cline{2-13}
& SP & MAE & RMSE & SP & MAE & RMSE & SP & MAE & RMSE & SP & MAE & RMSE & & \\ \hline\hline
\multirow{3}{*}{\shortstack{\textsc{Karma}\\(ours)}} 
& 100 & 0.67 & 0.90 & 100 & 0.70 & 0.96 & 100 & 0.61 & 0.83 & 100 & 0.68 & 0.76 & 0.67 & 0.86 \\ \cline{2-15}
& 199 & 0.42 & \textbf{0.64} & 199 & 0.60 & 0.89 & 171 & 0.47 & 0.71 & 171 & 0.42 & \textbf{0.63} & 0.48 & 0.72 \\ \cline{2-15}
& 300 & \textbf{0.40} & 0.65 & 300 & \textbf{0.51} & \textbf{0.85} & 300 & \textbf{0.45} & \textbf{0.69} & 300 & \textbf{0.41} & \textbf{0.63} & \textbf{0.44} & \textbf{0.70} \\ \hline
\textsc{DualStr} & 300 & 0.42 & 0.66 & 300 & 0.57 & 0.87 & 300 & 0.47 & \textbf{0.69} & 300 & 0.44 & 0.65 & 0.48 & 0.72 \\ \hline
Ref. \cite{wang2025exponential} & 301 & 0.45 & 0.66 & 301 & 0.61 & 0.95 & 301 & 0.52 & 0.81 & 301 & 0.54 & 0.75 & 0.53 & 0.79 \\ \hline
Ref. \cite{wang2023remaining} & 320 & 1.76 & 2.19 & 320 & 1.90 & 2.38 & 320 & 1.24 & 1.78 & 320 & 1.22 & 1.62 & 1.53 & 2.00 \\ \hline
Ref. \cite{li2023remaining} & 199 & 0.43 & 0.65 & 199 & 0.60 & 0.90 & 171 & 0.50 & 0.75 & 171 & 0.47 & 0.69 & 0.50 & 0.75 \\ \hline
Ref. \cite{zraibi2021remaining} & -- & -- & -- & 199 & 0.78 & 0.93 & 171 & 0.68 & 0.84 & -- & -- & -- & 0.73 & 0.89 \\ \hline
Ref. \cite{bao2025lightweight} & 100 & 0.71 & 0.93 & 100 & 1.17 & 1.65 & 100 & 0.74 & 0.97 & 100 & 0.72 & 1.01 & 0.84 & 1.14 \\ \hline\hline
\end{tabular}%
\label{tab:comp-rul}
\end{table*}

With different degradation patterns, the performance gap between \textsc{Karma} and the rest algorithms widens on CALCE batteries compared to NASA batteries. For the best alternative model, either CNN-LSTM or CNN-BiGRU in terms of MAE, the per-battery gap averages 0.0054, with individual gaps of 0.0074, 0.0065, 0.0045 and 0.0031 for the four batteries, respectively. For CNN-LSTM, the average MAPE rises from 0.672\% for NASA to 2.062\% for CALCE, and the average MAPE of \textsc{Karma} is still below 1\% for CALCE. The results show that \textsc{Karma}, integrating frequency-adaptive learning and knowledge-based modeling, generalizes better across different batteries and has greater practical values as comparison algorithms cannot adapt as well as \textsc{Karma}.

\paragraph{Computational Efficiency Analysis} 
\textsc{Karma} employs a dual-stream architecture for different signal frequencies alongside a knowledge-based model. Consequently, its system complexity and computational resource requirements are inherently higher than those of the comparison algorithms presented above. In the following, we analyze this increase in complexity and resource demand, with the expectation that the growth remains non-exponential to preserve practical feasibility for potential edge deployment. \textsc{Karma} comprises 236,801 parameters with a computational cost of 681,937 FLOPs, which are 69\% and 88\% higher than one of the best comparison algorithms, CNN-LSTM, respectively. These increases, however, are not exponential and remain within a reasonable range for deployment. More importantly, the performance gains achieved by \textsc{Karma} as presented above are substantial enough to offset the additional computational overhead. In addition, despite its enhanced architecture, \textsc{Karma} maintains a compact model size of 0.90 MB, which can be readily accommodated in many edge computing environments. The size difference compared with CNN-LSTM’s 0.53 MB model is marginal in practical terms and does not significantly increase the demand on upgrading computing facilities.


\setlength{\tabcolsep}{4pt} 
\begin{table*}[t]
\centering
\renewcommand{\arraystretch}{1.3}
\caption{The actual and predicted RUL for all the NASA and CALCE batteries with 95\% CI uncertainty quantification for different SPs.}
\begin{tabular}{c|cccc|cccc|cccc|cccc}
\hline\hline
\multirow{3}{*}{SP} & \multicolumn{16}{c}{NASA Dataset} \\ \cline{2-17} & \multicolumn{4}{c|}{B0005} & \multicolumn{4}{c|}{B0006} & \multicolumn{4}{c|}{B0007} & \multicolumn{4}{c}{B0018} \\ \cline{2-17} 
& Actual & Pred. & AE & 95\% CI & Actual & Pred. & AE & 95\% CI & Actual & Pred. & AE & 95\% CI & Actual & Pred. & AE & 95\% CI \\ \hline
50 & 74 & 74 & 0 & {[}66, 78{]} & 59 & 58 & 1 & {[}52, 61{]} & 96 & 97 & 1 & {[}90, 110{]} & 47 & 47 & 0 & {[}44, 51{]} \\ \hline
70 & 54 & 54 & 0 & {[}45, 58{]} & 39 & 38 & 1 & {[}36, 42{]} & 76 & 76 & 0 & {[}70, 89{]} & 27 & 27 & 0 & {[}24, 30{]} \\ \hline
90 & 34 & 34 & 0 & {[}26, 38{]} & 19 & 17 & 2 & {[}11, 20{]} & 56 & 56 & 0 & {[}50, 70{]} & 7 & 7 & 0 & {[}4, 10{]} \\ \hline\hline
\multirow{3}{*}{SP} & \multicolumn{16}{c}{CALCE Dataset} \\ \cline{2-17} & \multicolumn{4}{c|}{CS35} & \multicolumn{4}{c|}{CS36} & \multicolumn{4}{c|}{CS37} & \multicolumn{4}{c}{CS38} \\ \cline{2-17} 
& Actual & Pred. & AE & 95\% CI & Actual & Pred. & AE & 95\% CI & Actual & Pred. & AE & 95\% CI & Actual & Pred. & AE & 95\% CI \\ \hline
100 & 540 & 541 & 1 & {[}534, 549{]} & 545 & 539 & 6 & {[}508, 556{]} & 617 & 613 & 4 & {[}597, 652{]} & 654 & 659 & 5 & {[}650, 664{]} \\ \hline
200 & 440 & 440 & 0 & {[}433, 448{]} & 445 & 438 & 7 & {[}408, 456{]} & 517 & 518 & 1 & {[}501, 559{]} & 554 & 559 & 5 & {[}555, 561{]} \\ \hline
300 & 340 & 340 & 0 & {[}334, 349{]} & 345 & 341 & 4 & {[}308, 356{]} & 417 & 415 & 2 & {[}394, 453{]} & 454 & 456 & 2 & {[}454, 460{]} \\ \hline\hline
\end{tabular}%
\label{tab:uncertainty}
\end{table*}
\setlength{\tabcolsep}{6pt}

\subsubsection{RUL Prediction}
We now extend the analysis from 1-cycle capacity to RUL prediction. In \textsc{Karma}, the capacity predictions are used to fit the knowledge-based function cycle by cycle to find the cycle at which the battery reaches its EoL, and accordingly RUL can be calculated. We specifically consider the capacity prediction of the EoL cycle to quantify RUL prediction performance, e.g., small EoL capacity estimation errors lead to accurate RUL predictions. We compare \textsc{Karma} against several latest algorithms with diverse modeling strategies. First, we consider \textsc{Karma} without knowledge, i.e., our dual-stream frequency-adaptive model, denoted as \textsc{DualStr} for short. Such a comparison allows us to understand the potential of our data-driven model and the impact of knowledge guidance. Then, we consider hybrid ML models. One is CNN-LSTM-DNN \cite{zraibi2021remaining}, which combines convolutional feature extraction with sequential analysis, and is not frequency-adaptive. We also consider models largely driven by temporal patterns, e.g., TCN-GRU-DNN \cite{li2023remaining} and LTM-Net \cite{bao2025lightweight}. Furthermore, we consider the Exponential Transformer \cite{wang2025exponential} and a temporal-differential guided model \cite{wang2023remaining}. Performance is assessed with MAE and RMSE and SP is specified for different tests. We present the experimental results in Table \ref{tab:comp-rul}. For fair comparison, the results of the comparison algorithms are based on references, to avoid potential performance variations in different implementations. Note that the SP of the different algorithms/references vary. We use small SP, indicating more challenging RUL prediction with less observed battery health data, for our proposed \textsc{Karma}.

\paragraph{NASA Data}
Seen from Table \ref{tab:comp-rul}, \textsc{Karma} achieves the lowest errors across all four batteries when compared with other algorithms. This indicates \textsc{Karma}'s strengths in both prediction accuracy and reliability. Numerically, competing methods show significantly higher errors, with the same or larger SP. For example, the transformer model \cite{wang2025exponential} achieves MAE values in the range of 0.0052 to 0.0067 and RMSE values between 0.0075 and 0.0125. The temporal and differential guided algorithm \cite{wang2023remaining} performs worse, with up to 0.0213 MAE and 0.0286 RMSE. These results show that, while existing algorithms can capture general degradation trends, they cannot deliver precise EoL capacity estimates for NASA data. This again highlights \textsc{Karma}'s advantage in its synergy between data- and knowledge-driven models which result in consistently lower errors across different metrics.

We perform a direct comparison between \textsc{Karma} and our pure data-driven model \textsc{DualStr}. The results show that the impact of knowledge is significant. Without knowledge, \textsc{DualStr} only manages to deliver the same level of performance for B0005 in terms of MAE, where \textsc{DualStr}'s error is actually slightly higher when we consider three decimal places. For the rest three batteries, the MAE increases are 4.1\%, 5.1\% and 0.4\%, and the RMSE increases are more significantly, i.e., 8.1\%, 3.7\% and 13.3\%, respectively. The results suggest that our knowledge-based model well reflects the degradation trend and using the data-driven model to calibrate the knowledge-based model reduces noise sensitivity and enhances prediction stability compared to the pure data-driven model.

Even without knowledge, our data-driven \textsc{DualStr} delivers performance superior to the comparison algorithms. For example, at SP 60, its average MAE is 0.0055 only, compared to 0.0094, 71\% higher, for the recent transformer model \cite{wang2025exponential}. \textsc{DualStr}'s performance advantage compared to TCN-GRU-DNN \cite{li2023remaining} and CNN-LSTM-DNN \cite{zraibi2021remaining} is even more significant, with average MAEs of 0.0141 and 0.0180, which are 2.6x and 3.3x higher, respectively. The results show that our signal decomposition and frequency adaptive learning with a dual-stream model is effective in modelling degradation, and \textsc{Karma}'s competitiveness is beyond knowledge.

\paragraph{CALCE Data}
Seen from Table \ref{tab:comp-rul}, \textsc{Karma} demonstrates the best performance consistently, same as our observations for NASA dataset. For example, the average MAE of the best-performing comparison algorithm \cite{li2023remaining} is 0.0050, and \textsc{Karma} reduces the error by 12\% to 0.0044. We considered SPs from 100 to 300, and the average MAE steadily decreases from 0.0067 to 0.0044, with a 34\% of error reduction. RMSE values follow the same pattern, with a 17\% error reduction. This trend shows \textsc{Karma}'s ability to progressively refine RUL predictions with more available cycles, which contribute to deriving more accurate parameters of the knowledge. 

\textsc{Karma} is also practical in terms of its early prediction performance. For example, with as little as 100 cycles of observed data, \textsc{Karma} achieves MAE values below 0.0070 across all CALCE batteries. In comparison, \cite{bao2025lightweight} with the same SP achieves only 0.0117 MAE for CS36, 67\% higher than \textsc{Karma}'s MAE. The SP is nearly doubled for \cite{zraibi2021remaining}, which however has 11\% higher MAE compared to \textsc{Karma} with SP 100. This suggests that \textsc{Karma}, although benefits from more cycles of data, can support decision-making at an earlier stage of battery life. This is valuable for predictive maintenance and safety-critical applications. Overall, the comprehensive results across both NASA and CALCE datasets establish \textsc{Karma} as a robust and reliable solution for RUL prediction across diverse battery technologies and operating environments.

\begin{figure*}[t]
    \centering
    \def \tmph{1.45in}
    \subfigure[NASA Dataset]{    \includegraphics[width=0.98\linewidth]{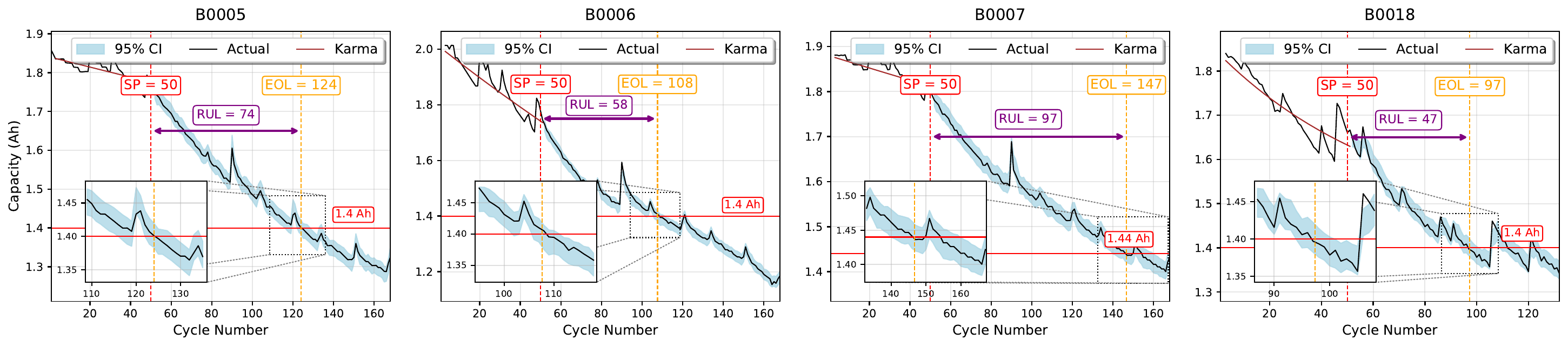}\label{fig:uncertainty-nasa}}
    \subfigure[CALCE Dataset]{\includegraphics[width=0.98\linewidth]{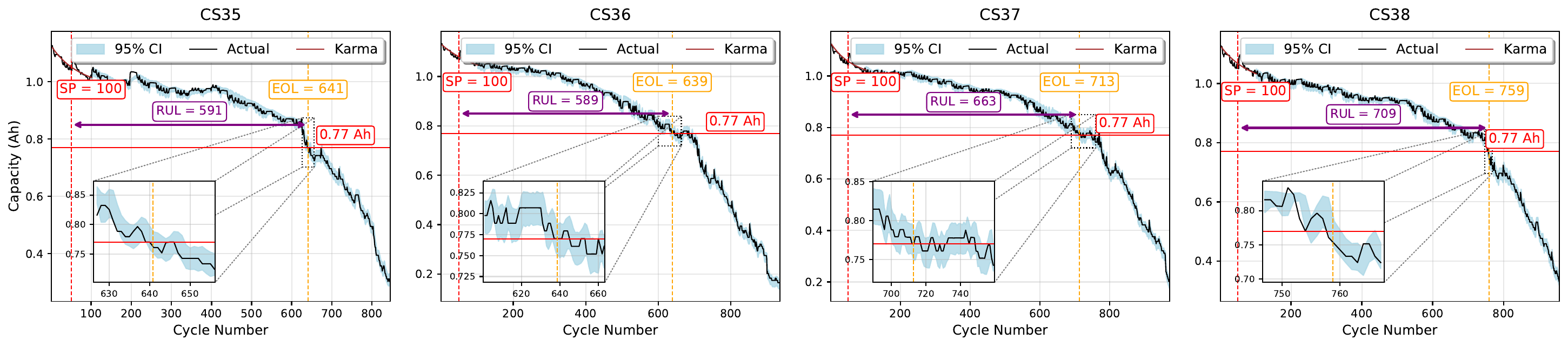}\label{fig:uncertainty-calce}}
    \caption{\textsc{Karma}'s battery capacity estimation with 95\% CI uncertainty quantification and RUL prediction for NASA batteries at SP 50 in Fig. \ref{fig:uncertainty-nasa} and CALCE batteries at SP 100 in Fig. \ref{fig:uncertainty-calce}.}
    \label{fig:uncertainty}
\end{figure*}

\subsection{Uncertainty Analysis}
Accurate estimation of battery health is important; however, point estimates alone, even accurate, may be misleading in practice. Decision-making in industry applications often requires not only high prediction accuracy but also reliable quantification of prediction confidence. \textsc{Karma} naturally supports uncertainty analysis through its PF-based modelling, where statistical information like confidence intervals (CIs) can be calculated based on a set of different particles. This feature allows end-users to assess the reliability of predictions and adapt operational strategies according to the associated uncertainty. In this part, we show the actual and predicted EoL cycles with uncertainty quantification in Table \ref{tab:uncertainty} for NASA and CALCE datasets, respectively. Several representative capacity trajectories and prediction results are shown in Figs. \ref{fig:uncertainty-nasa} and \ref{fig:uncertainty-calce}. We can observe \textsc{Karma}'s competitive prediction accuracy, with absolute errors (AEs) consistently ranging from 0 to 2 cycles for NASA batteries. For certain test cases, e.g., B0005 with SPs 50, 70 and 90, \textsc{Karma} has perfect predictions with zero error in predicting the cycle of EoL. CIs reveal the important differences in uncertainty. For example, B0007 exhibits the widest CIs with 20 cycles between the upper and lower limits of the CI, indicating a relatively high level of prediction uncertainty, and the CIs of B0018 remain narrow, e.g., width $\leq 7$, which implies good predictive reliability. For CALCE batteries, we observe significantly larger values of the actual and predicted EoL cycles compared to NASA batteries, mainly because of the long lifetime of CALCE batteries. Across the CALCE batteries, the AEs range from 0 to 7 cycles, where \textsc{Karma} has perfect estimations for CS35 with SPs 200 and 300. Early prediction for CS36 at SP 100 is challenging where \textsc{Karma} has an error of 6 cycles. The prediction uncertainty for CS36 is also high, with CIs of width 48 for several SPs.

A general trend is that CIs do not widen as the SP approaches the cycle of EoL, because of the benefit of having more observed degradation data. This trend is evident in some test cases, e.g., CS38 where the CI width shrinks from 14 to 6 cycles from SPs from 100 to 300. However, this observation does not hold for all test cases. For example, B0007 and CS36 maintain relatively wide CIs even at later SPs. This can be attributed to the irregularities and noises in the entire degradation trajectories, which bring challenges for different models to reduce uncertainty effectively despite more data until SP being available. From a practical perspective, we can see the uncertainty quantification is meaningful even at the early stage of prediction. For example, \textsc{Karma} performs competitively with narrow enough CIs at SPs 50 and 100 for NASA and CALCE datasets, respectively. The integration of accuracy and uncertainty ensures that a model does not overstate its performance, allowing optimal decisions for predictive maintenance and safety monitoring.

\begin{table*}[t]
\centering
\renewcommand{\arraystretch}{1.3}
\caption{Performance of our dual-stream model across different forecast horizons in terms of MAE ($\times 10^{-2}$), RMSE ($\times 10^{-2}$), MAPE (\%), and the overall average MAPE (\%). Longer horizons present greater challenges with higher errors.
}
\begin{tabular}{c|ccc|ccc|ccc|ccc|c}
\hline\hline
\multirow{3}{*}{\shortstack{Forecast\\Horizon}} & \multicolumn{13}{c}{NASA Dataset} \\ \cline{2-14}
& \multicolumn{3}{c|}{B0005} & \multicolumn{3}{c|}{B0006} & \multicolumn{3}{c|}{B0007} & \multicolumn{3}{c|}{B0018} & \multirow{2}{*}{\shortstack{Average\\ MAPE}} \\\cline{2-13}
& MAE & RMSE & MAPE & MAE & RMSE & MAPE & MAE & RMSE & MAPE & MAE & RMSE & MAPE &  \\ \hline\hline
1  & 0.323 & 0.45  & 0.227 & 0.534 & 0.666 & 0.382 & 0.236 & 0.352 & 0.152 & 0.536 & 0.742 & 0.374 & 0.284 \\\hline
5  & 0.808 & 1.235 & 0.561 & 1.235 & 1.895 & 0.907 & 0.691 & 1.144 & 0.450 & 1.811 & 2.455 & 1.264 & 0.796 \\ \hline
10 & 1.262 & 1.672 & 0.871 & 2.418 & 3.656 & 1.731 & 0.969 & 1.582 & 0.627 & 2.499 & 3.068 & 1.758 & 1.247 \\ \hline
15 & 1.284 & 1.630 & 0.896 & 3.476 & 4.619 & 2.499 & 0.922 & 1.452 & 0.593 & 2.841 & 3.569 & 2.001 & 1.497 \\
\hline\hline
\multirow{3}{*}{\shortstack{Forecast\\Horizon}} & \multicolumn{13}{c}{CALCE Dataset} \\ \cline{2-14}  
& \multicolumn{3}{c|}{CS35} & \multicolumn{3}{c|}{CS36} & \multicolumn{3}{c|}{CS37} & \multicolumn{3}{c|}{CS38} & \multirow{2}{*}{\shortstack{Average\\ MAPE}} \\\cline{2-13}
& MAE & RMSE & MAPE & MAE & RMSE & MAPE & MAE & RMSE & MAPE & MAE & RMSE & MAPE &  \\ \hline\hline
1  & 0.415 & 0.656 & 0.613 & 0.569 & 0.867 & 1.073 & 0.465 & 0.694 & 0.724 & 0.439 & 0.653 & 0.615 & 0.756 \\\hline
5  & 0.919 & 1.334 & 1.391 & 1.127 & 1.495 & 2.150 & 0.909 & 1.382 & 1.626 & 0.817 & 1.133 & 1.158 & 1.581 \\\hline
10 & 1.325 & 1.898 & 2.058 & 1.527 & 2.009 & 2.974 & 1.097 & 1.473 & 1.825 & 1.084 & 1.444 & 1.478 & 2.084 \\\hline
15 & 1.507 & 2.120 & 2.302 & 1.701 & 2.247 & 3.506 & 1.405 & 1.899 & 2.465 & 1.178 & 1.578 & 1.617 & 2.473 \\
\hline\hline
\end{tabular}
\label{tab:multi-step}
\end{table*}

\subsection{Forecast Horizon Sensitivity Analysis}
\textsc{DualStr} is the data-driven model we developed for \textsc{Karma}. It follows a dual-stream architecture which integrates CNN and LSTM for one frequency band and BiGRU for the other band. While it is part of \textsc{Karma}, \textsc{DualStr} alone can also model degradation dynamics and predict battery health in the future cycles, e.g., in a purely data-driven manner. In this part, we consider \textsc{DualStr}'s performance for different forecast horizons, including 1-, 5-, 10- and 15-cycle horizons, where \textsc{DualStr} is tasked to predict progressively further into the future. Table \ref{tab:multi-step} present the results for both NASA and CALCE datasets across different horizons. 

The results confirm that \textsc{DualStr} predicts most reliably in the near term, with accuracy gradually degrading as the forecast horizon extends. For 1-cycle prediction, the errors are consistently low across both NASA and CALCE datasets, with errors almost negligible in some test cases, e.g., 0.002 MAE for B0007. For medium horizons, the errors remain within acceptable margins, e.g., 0.8\% MAPE for 5-cycle prediction, and the data-driven model retains practical usefulness. When we stretch the horizon beyond 10 cycles, the average MAPE exceeds one percent, e.g., 2.5\% for B0006 for 15-cycle prediction. This trend is expected for data-driven models, since each iterative forecast is influenced by prior inaccuracies. The results also highlights the importance of regulating the degradation trend modelling with knowledge to avoid performance drift for the solutions that are purely data-driven. 

We also notice that the performance differs significantly among tested batteries. For instance, the errors for B0007 and CS38 are generally low across horizons, meaning that the degradation patterns are easier for the model to capture. In contrast, the errors increase significantly for some batteries with different prediction horizons, e.g., 0.0053 MAE to 0.0348 for B0006 with 6.5x increase. This happens for batteries with more complex degradation patterns such as noisier capacity trajectories. These differences align with an intuitive understanding that data or a test case itself has a big impact on model performance, and the model is not the only determine factor. Designing a model which can adapt to complex behaviors across battery chemistry and usage profiles is highly important and this is one of the motivations of the developing the knowledge-guided \textsc{Karma}.

\balance
\section{Conclusion}
\label{sec:conclusion}
This paper presents \textsc{Karma}, a novel hybrid model that combines frequency-adaptive deep learning with knowledge-regulated modeling for battery health prognostics. We decompose raw battery signals into low- and high-frequency IMFs and process them with specialized ML streams, which effectively capture both long-term degradation trends and short-term fluctuations, respectively. The integration of an empirical knowledge, i.e., modelling the degradation trend as a double exponential function, ensures that battery health predictions remain knowledge-guided and are accurate and robust. PF is employed to optimize the parameters of the double exponential function and the set of particles allow quantifying prediction uncertainties. Comprehensive experiments on two mainstream datasets, for NASA and CALCE batteries, demonstrate the superior performance of \textsc{Karma} over state-of-the-art comparison algorithms. On average, \textsc{Karma} reduces the MAE of the comparison algorithms by 50.6\% and 32.6\% for NASA and CALCE datasets, respectively. Results for early-cycle prediction, uncertainty quantification and forecast horizon sensitivity further confirm \textsc{Karma}'s robustness and generalizability. Overall, \textsc{Karma} is a novel and practical solution for reliable and trustworthy battery management systems.

In the future, \textsc{Karma} can be customized and applied for the batteries of different battery types, e.g., solid-state and sodium-ion batteries. We also would like to optimize \textsc{Karma} for real-world deployment and develop lightweight versions for edge intelligence. Furthermore, we believe integrating knowledge and data-driven models has broad potential in other industry applications like machinery predictive maintenance. 

\ifCLASSOPTIONcaptionsoff
  \newpage
\fi



\bibliographystyle{IEEEtran}
\bibliography{IEEEabrv,ref_IEEE}

\begin{thebibliography}{10}
\providecommand{\url}[1]{#1}
\csname url@samestyle\endcsname
\providecommand{\newblock}{\relax}
\providecommand{\bibinfo}[2]{#2}
\providecommand{\BIBentrySTDinterwordspacing}{\spaceskip=0pt\relax}
\providecommand{\BIBentryALTinterwordstretchfactor}{4}
\providecommand{\BIBentryALTinterwordspacing}{\spaceskip=\fontdimen2\font plus
\BIBentryALTinterwordstretchfactor\fontdimen3\font minus \fontdimen4\font\relax}
\providecommand{\BIBforeignlanguage}[2]{{%
\expandafter\ifx\csname l@#1\endcsname\relax
\typeout{** WARNING: IEEEtran.bst: No hyphenation pattern has been}%
\typeout{** loaded for the language `#1'. Using the pattern for}%
\typeout{** the default language instead.}%
\else
\language=\csname l@#1\endcsname
\fi
#2}}
\providecommand{\BIBdecl}{\relax}
\BIBdecl

\bibitem{zhang2024role}
H.~Zhang, D.~Niyato, W.~Zhang \emph{et~al.}, ``The role of generative artificial intelligence in internet of electric vehicles,'' \emph{IEEE Internet of Things Journal}, 2024.

\bibitem{qi2024joint}
L.~Qi, B.~Wu, X.~Chen \emph{et~al.}, ``Joint optimization of internet of things and smart grid for energy generation, battery (dis)charging, and information delivery,'' \emph{IEEE Internet of Things Journal}, vol.~11, no.~12, pp. 21\,647--21\,658, 2024.

\bibitem{fleischmann2023battery}
J.~Fleischmann, M.~Hanicke, E.~Horetsky \emph{et~al.}, ``Battery 2030: Resilient, sustainable, and circular,'' \emph{McKinsey \& Company}, pp. 2--18, 2023.

\bibitem{Zhao2024Practical}
Y.~Zhao, W.~Zhang, Q.~Yan \emph{et~al.}, ``Practical battery health monitoring using uncertainty-aware bayesian neural network,'' in \emph{2024 IEEE 100th Vehicular Technology Conference (VTC2024-Fall)}, 2024, pp. 1--6.

\bibitem{zhang2018long}
Y.~Zhang, R.~Xiong, H.~He \emph{et~al.}, ``Long short-term memory recurrent neural network for remaining useful life prediction of lithium-ion batteries,'' \emph{IEEE Transactions on Vehicular Technology}, vol.~67, no.~7, pp. 5695--5705, 2018.

\bibitem{li2023remaining}
L.~Li, Y.~Li, R.~Mao \emph{et~al.}, ``Remaining useful life prediction for lithium-ion batteries with a hybrid model based on tcn-gru-dnn and dual attention mechanism,'' \emph{IEEE Transactions on Transportation Electrification}, vol.~9, no.~3, pp. 4726--4740, 2023.

\bibitem{zraibi2021remaining}
B.~Zraibi, C.~Okar, H.~Chaoui \emph{et~al.}, ``Remaining useful life assessment for lithium-ion batteries using cnn-lstm-dnn hybrid method,'' \emph{IEEE Transactions on Vehicular Technology}, vol.~70, no.~5, pp. 4252--4261, 2021.

\bibitem{anh2024prediction}
V.~Q. Anh, N.~D. Tuyen, G.~Fujita \emph{et~al.}, ``Prediction of state-of-health and remaining-useful-life of battery based on hybrid neural network model,'' \emph{IEEE Access}, 2024.

\bibitem{wang2023remaining}
T.~Wang, Z.~Ma, and S.~Zou, ``Remaining useful life prediction of lithium-ion batteries: A temporal and differential guided dual attention neural network,'' \emph{IEEE Transactions on Energy Conversion}, vol.~39, no.~1, pp. 757--771, 2023.

\bibitem{wang2025exponential}
C.~Wang, Z.~Bao, H.~Lin \emph{et~al.}, ``An exponential transformer for learning interpretable temporal information in remaining useful life prediction of lithium-ion battery,'' \emph{IEEE Transactions on Transportation Electrification}, 2025.

\bibitem{park2025detailed}
J.~Park, G.~H. Lee, J.~Kim \emph{et~al.}, ``Detailed architectural design of a multi-head self-attention model for lithium-ion battery capacity forecasting,'' \emph{IEEE Access}, 2025.

\bibitem{yao2025remaining}
X.~Yao, K.~Su, H.~Zhang \emph{et~al.}, ``Remaining useful life prediction for lithium-ion batteries in highway electromechanical equipment based on feature-encoded lstm-cnn network,'' \emph{Energy}, vol. 323, p. 135719, 2025.

\bibitem{10253731}
J.~Zhang, J.~Tian, Y.~Jiang \emph{et~al.}, ``Sagpcn: Self-attention graph pooling convolutional network for battery state of health estimation,'' in \emph{2023 IEEE 3rd International Conference on Industrial Electronics for Sustainable Energy Systems (IESES)}, 2023, pp. 1--6.

\bibitem{9040661}
K.~Liu, Y.~Shang, Q.~Ouyang \emph{et~al.}, ``A data-driven approach with uncertainty quantification for predicting future capacities and remaining useful life of lithium-ion battery,'' \emph{IEEE Transactions on Industrial Electronics}, vol.~68, no.~4, pp. 3170--3180, 2021.

\bibitem{duan2025lithium}
C.~Duan, H.~Cao, F.~Liu \emph{et~al.}, ``Lithium-ion batteries remaining useful life prediction using a parallel bilstm-mha neural network based on a ceemdan module,'' \emph{The International Journal of Advanced Manufacturing Technology}, vol. 137, no.~7, pp. 3359--3386, 2025.

\bibitem{yuan2024improved}
H.~Yuan, J.~Bi, S.~Li \emph{et~al.}, ``An improved lstm-based prediction approach for resources and workload in large-scale data centers,'' \emph{IEEE Internet of Things Journal}, vol.~11, no.~12, pp. 22\,816--22\,829, 2024.

\bibitem{6655981}
K.~Dragomiretskiy and D.~Zosso, ``Variational mode decomposition,'' \emph{IEEE Transactions on Signal Processing}, vol.~62, no.~3, pp. 531--544, 2014.

\bibitem{9758685}
G.~Ding, W.~Wang, and T.~Zhu, ``Remaining useful life prediction for lithium-ion batteries based on cs-vmd and gru,'' \emph{IEEE Access}, vol.~10, pp. 89\,402--89\,413, 2022.

\bibitem{CHEN2024113388}
Y.~Chen, X.~Huang, Y.~He \emph{et~al.}, ``Edge–cloud collaborative estimation lithium-ion battery soh based on mewoa-vmd and transformer,'' \emph{Journal of Energy Storage}, vol.~99, p. 113388, 2024.

\bibitem{bao2025lightweight}
Z.~Bao, T.~Luo, M.~Gao \emph{et~al.}, ``A lightweight and term-arbitrary memory network for remaining useful life prediction of li-ion battery,'' \emph{IEEE Transactions on Instrumentation and Measurement}, 2025.

\bibitem{liang2024hybrid}
J.~Liang, H.~Liu, and N.-C. Xiao, ``A hybrid approach based on deep neural network and double exponential model for remaining useful life prediction,'' \emph{Expert Systems with Applications}, vol. 249, p. 123563, 2024.

\bibitem{xie2020prognostic}
R.~Xie, R.~Ma, S.~Pu \emph{et~al.}, ``Prognostic for fuel cell based on particle filter and recurrent neural network fusion structure,'' \emph{Energy and AI}, vol.~2, p. 100017, 2020.

\bibitem{zhang2023data}
J.~Zhang, C.~Huang, M.-Y. Chow \emph{et~al.}, ``A data-model interactive remaining useful life prediction approach of lithium-ion batteries based on pf-bigru-tsam,'' \emph{IEEE Transactions on Industrial Informatics}, vol.~20, no.~2, pp. 1144--1154, 2023.

\bibitem{lu2025remaining}
Y.~Lu, Y.~Shi, Y.~Liu \emph{et~al.}, ``Remaining useful lifetime prediction of lithium-ion batteries based on fragment data and trend identification,'' \emph{IEEE Transactions on Industrial Informatics}, 2025.

\bibitem{chen2024hybrid}
Z.~Chen, Z.~Wang, W.~Wu \emph{et~al.}, ``A hybrid battery degradation model combining arrhenius equation and neural network for capacity prediction under time-varying operating conditions,'' \emph{Reliability Engineering \& System Safety}, vol. 252, p. 110471, 2024.

\bibitem{pamshetti2025optimal}
V.~B. Pamshetti, W.~Zhang, K.~J. Tseng \emph{et~al.}, ``Optimal signal decomposition-based multi-stage learning for battery health estimation,'' in \emph{2025 IEEE Intelligent Vehicles Symposium (IV)}, 2025, pp. 2409--2414.

\bibitem{li2023hybrid}
Y.~Li, L.~Li, R.~Mao \emph{et~al.}, ``Hybrid data-driven approach for predicting the remaining useful life of lithium-ion batteries,'' \emph{IEEE Transactions on Transportation Electrification}, vol.~10, no.~2, pp. 2789--2805, 2023.

\bibitem{saha2007battery}
B.~Saha and K.~Goebel, ``Battery data set,'' NASA AMES Prognostics Data Repository, USA, 2025.

\bibitem{calce2017lithium}
\BIBentryALTinterwordspacing
CALCE, ``Lithium-ion battery experimental data,'' Online, 2025, accessed: Jun. 5, 2025. [Online]. Available: \url{https://web.calce.umd.edu/batteries/data.htm}
\BIBentrySTDinterwordspacing

\end{thebibliography}

%

\end{document}